\documentclass{article}
\usepackage{ijcai17}

\usepackage{times}
\usepackage{physics}
\usepackage{graphicx}
\usepackage{subcaption}

\title{Reconciling Irrational Human Behavior with AI based Decision Making: A Quantum Probabilistic Approach }
\author{Sagar Uprety\\
The Open University, UK\\
sagar.uprety@open.ac.uk
\And 
Dawei Song\\
The Open University, UK\\
Beijing Institute of Technology, China\\
dawei.song@open.ac.uk
}

\begin{document}

\maketitle

\begin{abstract}
There are many examples of human decision making which cannot be modeled by classical probabilistic and logic models, on which the current AI systems are based. Hence the need for a modeling framework which can enable intelligent systems to detect and predict cognitive biases in human decisions to facilitate better human-agent interaction. We give a few examples of irrational behavior and use a generalized probabilistic model inspired by the mathematical framework of Quantum Theory to model and explain such behavior. 
\end{abstract}

\section{Introduction}
Decades of research by cognitive scientists have shown that in some cases human judgment under uncertainty violates the classical(Bayesian) Probability theory and other logic models~\cite{kahneman-judment-book,Tversky1992-disjunction-effect}. In a famous experiment, \cite{Tversky1983-conjunction-fllacy} presented participants with the following text:
\\
\\
\footnotesize
\textbf{Linda is 31 years old, single, outspoken and very bright. She majored in philosophy. As a student, she was deeply concerned with the issues of discrimination and social justice, and also participated in anti-nuclear demonstrations. Which is more probable:
(a) Linda is a bank teller
(b) Linda is active in the feminist movement and is a bank teller}
\\
\\
\normalsize
The participants consistently rated the probability of event (b) as more than that of (a). This violates the axioms of probability theory, according to which the probability of conjunction of two events is always less than that of any of the single events. In the set theoretical formalism of probability, of the sample space of all possible Lindas who are Bank Tellers, only a subset of it will be both Bank Teller and Feminist.

These findings, termed as the Conjunction Fallacy, have been investigated a lot since then~\cite{Sides2002}. Experiments have been conducted with various kinds of stories, even using words like "betting" instead of "Probability", indicating that this judgment error is not due to ignorance or misunderstanding of the concept of Probability. There is also another example of similar behavior called the Disjunction Fallacy, where humans rate the probability of disjunction as less than that of individual events. It is concluded that the classical probability Theory cannot explain such judgments. 

Another paradoxical finding from the works of Tversky is that similarity judgments by humans violate metric axioms. In some cases, the similarity of A and B is not the same as similarity of B and A. As an example, the similarity of Korea(North Korea) to China was judged greater than the similarity of China to Korea~\cite{Tversky1977-similarity}. The explanation proposed by Tversky was that most of the features associated with Korea are similar to China. So Korea appears more similar to China. However, China has many other features associated with it. One has more knowledge about China than Korea, while judging Sim(China, Korea). Therefore it does not appear as similar as Sim(Korea, China). Similarity between two objects is a function of distance between points in a multidimensional space, the objects being represented by the points. Thus it should not depend upon the order in which the objects are considered. So this is another instance where human decision making does not conform to the existing methods of modeling. 

Different explanations and models have been proposed for the judgment fallacies described above~\cite{Tversky1983-conjunction-fllacy,Tversky1977-similarity,Nosofsky1984,Nosofsky1991,Krumhansl1978,Krumhansl1988,Ashby1988}. We now present a generalized probabilistic model which can incorporate and explain the above judgment fallacies.

\section{Quantum Probabilistic Modeling}

\begin{figure*}[t!]
    \begin{subfigure}[t]{0.24\textwidth}
        \includegraphics[width=\textwidth]{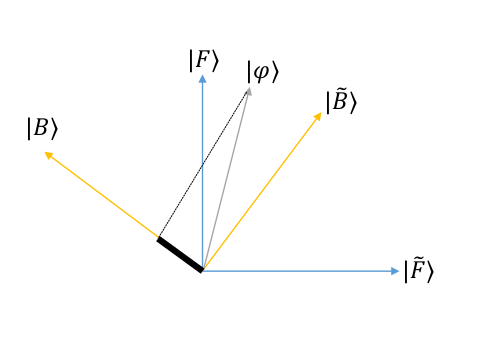}
        \caption{Figure 1.a}
    \end{subfigure}
    \begin{subfigure}[t]{0.24\textwidth}
        \includegraphics[width=\textwidth]{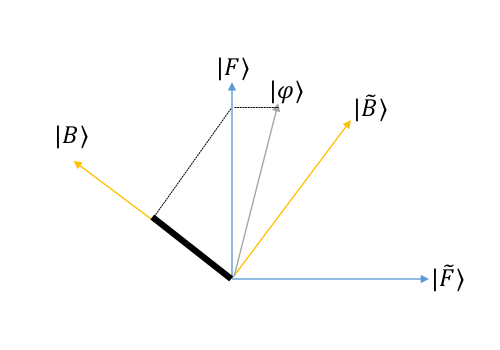}
        \caption{Figure 1.b}
     \end{subfigure}
      \begin{subfigure}[t]{0.24\textwidth}
        \includegraphics[width=\textwidth]{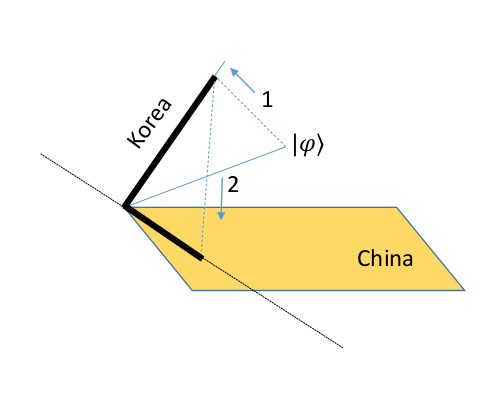}
        \caption{Figure 1.c}
    \end{subfigure}
    \begin{subfigure}[t]{0.24\textwidth}
        \includegraphics[width=\textwidth]{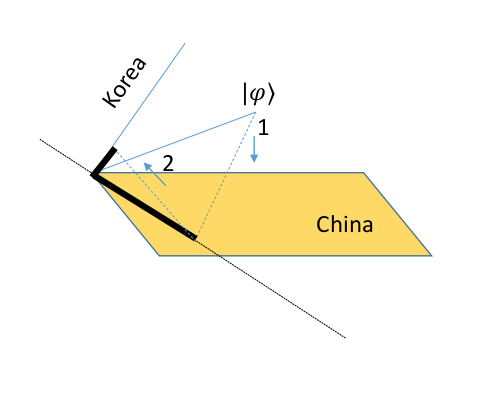}
        \caption{Figure 1.d}
    \end{subfigure}
\end{figure*}

Quantum Theory was developed to explain the counter-intuitive behavior of microscopic particles, which could not be modeled using standard probability theories. It was axiomatically organized by von Neumann~\cite{9780691028934}, which enables it to be used as an abstract mathematical framework independent of Physics. In the standard probability theory, events are defined as subsets of a sample space of all possible events. In the Quantum logic framework, the sample space is a finite or infinite dimensional Hilbert space, which is an abstract Vector Space with inner products. Each event is represented as a subspace of the Hilbert space. For example, consider the event "Linda is active in the feminist movement". In a two dimensional Hilbert space, let $F$ be the vector(a one dimensional subspace) denoting this event. We rather denote it as $\ket{F}$, to be consistent with the Dirac notation of Quantum Theory. The negation of this event, that Linda is not active in the feminist movement is given by an orthogonal vector, denoted as $\ket{\widetilde{F}}$. Together, these two vectors span the two dimensional Hilbert space, thus forming an orthogonal basis. We have another event "Linda is a bank teller", $\ket{B}$. Now this event is not mutually exclusive to $\ket{F}$ or $\ket{\widetilde{F}}$, nor is same as them. So we denote it in the same Hilbert space as a separate vector. $\ket{B}$ and $\ket{\widetilde{B}}$ form another orthonormal basis of the Hilbert space. The Quantum equivalent of the probability distribution function - which assigns classical probabilities to each event, is an abstract state vector. The probability of an event is calculated by projecting the state vector onto the event subspace and taking the square of the projection obtained. The closer an event subspace is to the state vector, the larger the projection, and hence larger the probability. The essential difference between Quantum and classical probabilities lies in the concept of incompatible events. It is not possible to specify a joint probability distribution for incompatible events. Being certain about the outcome of one event induces an uncertain state regarding the outcomes of other events. In terms of cognition, incompatible events means that a cognitive agent cannot think about two events at the same time, thus assesses them one after the other. Incompatibility induces a sequence of judgments, instead of a joint distribution. For compatible events, the Quantum framework gives the same results as the classical one.  For the Quantum probabilistic modeling of the Conjunction Fallacy~\cite{Busemeyer2011-conjunction}, consider the Hilbert space in Figure 1. The state vector $\ket{\psi}$ represents user's cognitive state prior to evaluating the two questions posed about Linda in the previous section. Note that $\ket{\psi}$ is closer to $\ket{F}$ and almost orthogonal to $\ket{B}$, indicating that for the user, the probability that Linda is a feminist is high and that Linda is a bank teller(option (a) in the problem described above) is low. As the two events described in option (b) are represented as incompatible, the user cannot consider their joint probability and evaluates them sequentially. We, therefore, first project the state $\ket{\psi}$ onto $\ket{F}$ and then onto $\ket{B}$(Figure 1.b). This final projection is larger than the direct projection from $\ket{\psi}$ to $\ket{B}$(Figure 1.b). For this alignment of vectors, the Quantum model explains the Conjunction Fallacy.

For the explanation of asymmetry in similarity judgment, \cite{Pothos2011AQP} propose to model the distinct features of concepts as different subspaces. So concepts with higher number of features are represented as subspaces of higher dimensionality. Consider a simplified example of a three dimensional Hilbert space where the concept China is associated with a two dimensional subspace, and Korea is associated with a one-dimensional subspace. The initial cognitive state $\ket{\psi}$ is uniformly suspended between the two subspaces. For Sim(Korea, China), it is first projected onto the subspace for Korea and then onto the subspace for China(Figure 2.a) The order of projections is reversed for Sim(China, Korea). As can be seen(Figure 2.b), the final projection(Projection 2) is larger in the case of Sim(Korea, China). The geometrical reason behind this is that for the Sim(Korea, China) case, the last projection is to a higher dimensional subspace, which preserves a larger portion of the vector than a projection to a lower dimensional subspace. This also intuitively explains the fact that since China has more features than Korea, it is easier to think of those features which are similar to Korea(form of government, etc.), when evaluating Sim(Korea, China). 

\section{Conclusion and Future Work}
We showed some examples of human judgments which appear inexplicable by classical probability theories. Quantum theory provides a generalized, geometric theory of probability which provides parameter free modeling of many of such examples. It also incorporates the classical probability theory as a special case, where all events are compatible with each other. The Quantum framework is being increasingly applied to cognitive science under the field of Quantum Cognition~\cite{Busemeyer:2012:QMC:2385442}. From the systems side, it is being applied to Information Retrieval and Natural Language Processing~\cite{Rijsbergen:2004:GIR:993731,Sordoni2013,Bruza2011,Aerts2018}. The enigmatic Quantum principles of Superposition, Interference, Entanglement and Contextuality are being investigated and applied in all of these areas. What we propose in this paper is the need to enhance the capabilities of AI agents by modeling them using the Quantum Probabilistic framework. This will enable them to detect and predict irrational behavior, thus enabling them think more like humans. Not only that, according to ~\cite{Bruza2018} this will help humans put more trust in the autonomous systems. The first step would be to identify cases where irrational human behavior occurs and model them using Quantum Probability.

\bibliographystyle{named}
\bibliography{ijcai17}

\end{document}